\documentclass[]{llncs}

\usepackage{times}
\usepackage{xcolor}
\usepackage{soul}
\usepackage[utf8]{inputenc}
\usepackage[small]{caption}
\usepackage{booktabs}
\usepackage{algpseudocode}
\usepackage{algorithm}
\usepackage{float}
\usepackage{subcaption}
\usepackage{todonotes}
\usepackage{array}
\usepackage{makecell}
\usepackage{paralist}

\newcommand{\vbsine}{vb-union-SInE\ }

\renewcommand{\phi}{\varphi}
\algrenewcommand{\algorithmicrequire}{\textbf{Input:}}
\algrenewcommand{\algorithmicensure}{\textbf{Output:}}

\usepackage{tikz}
\usepackage{environ}

\usetikzlibrary{positioning}
\usetikzlibrary{patterns}

\usetikzlibrary{arrows.meta}
\tikzset{%
  >={Latex[width=2mm,length=2mm]},
            base/.style = {rectangle, rounded corners, draw=black,
                           minimum width=3.5cm, minimum height=0.7cm,
                           text centered, font=\sffamily},
  activityStarts/.style = {base, fill=gray!30},
       startstop/.style = {base, fill=red!30},
    activityRuns/.style = {base, fill=green!30},
    formula/.style = {base, minimum width = 1cm, draw=white},
    box/.style={%
    minimum width=4mm,%
    minimum height=2mm,%
    align=center},
        bla/.style = {base, minimum width = 0.2cm, fill=green!30, draw=white},
         process/.style = {base, minimum width=2.2cm, fill=orange!15,
                           font=\ttfamily},
}
\usetikzlibrary{shapes.geometric} 
\usetikzlibrary{arrows,backgrounds,calc,hobby,positioning}
\ExplSyntaxOn
\cs_if_exist:NF \prg_stepwise_function:nnnN { \cs_gset_eq:NN \prg_stepwise_function:nnnN \int_step_function:nnnN }
\cs_if_exist:NF \prg_stepwise_inline:nnnn { \cs_gset_eq:NN \prg_stepwise_inline:nnnn \int_step_inline:nnnn }
\ExplSyntaxOff
\tikzset{vertex style/.style={
    draw=#1,
    thick,
    text=black,
    ellipse,
    minimum width=0.5cm,
    minimum height=0.25cm,
    font=\footnotesize,
    outer sep=1pt,
  }, 
  text style/.style={
    sloped,
    text=black,
    font=\footnotesize,
    above
  }
}

\usepackage{graphicx}
\usepackage{amssymb}
\usepackage{amsmath}
\usepackage{adjustbox}
\usepackage{array}
\usepackage{mathtools}
\usepackage{tabularx}

\usepackage{caption}
\usepackage{sidecap}
\newcounter{rowcount}
\usepackage{caption}
\usepackage{tikz}
\usepackage{pgfplots}
\usepackage{rotating}
\usepackage[long]{optional}

\usepackage{multirow}

\pgfplotsset{compat=1.11,
        /pgfplots/ybar legend/.style={
        /pgfplots/legend image code/.code={%
        \draw[##1,/tikz/.cd,bar width=3pt,yshift=-0.2em,bar shift=0pt]
                plot coordinates {(0cm,0.8em)};},
},
}

\usepackage{filecontents}

\usepackage{etoolbox}
\usepackage{color}

\usepackage{hyperref}

\usepackage[shortlabels,inline]{enumitem}
\newcommand{\simvecs}{\mathit{simvecs}}
\newcommand{\simwords}{\mathit{simwords}}
\newcommand{\simtriggers}{\mathit{simtriggers}}

\newcommand{\tchanges}[1]{\textcolor{black}{#1}}

\newcommand{\repthanks}[1]{\textsuperscript{\ref{#1}}}
\makeatletter
\patchcmd{\maketitle}
  {\def\thanks}
  {\let\repthanks\repthanksunskip\def\thanks}
  {}{}
\patchcmd{\@maketitle}
  {\def\thanks}
  {\let\repthanks\@gobble\def\thanks}
  {}{}
\newcommand\repthanksunskip[1]{\unskip{}}

\makeatother

\title{Selection Strategies for Commonsense Knowledge}
\author{Claudia Schon\inst{1,2} \thanks{Work supported by DFG grant CoRg -- Cognitive Reasoning.} \\
\small{schon@uni-koblenz.de  } }
\date{} 
\institute{Institute for Web Science and Technologies, Universität Koblenz-Landau, Germany
\and
Institut für Informatik, Johannes Gutenberg-Universität Mainz, Germany}

\begin{document}

\maketitle

\begin{abstract} Selection strategies are broadly used in first-order logic theorem proving to select those parts of a large knowledge base that are necessary to proof a theorem at hand. Usually, these selection strategies do not take the meaning of symbol names into account. In knowledge bases with commonsense knowledge, symbol names are usually chosen to have a meaning and this meaning provides valuable information for selection strategies. 
We introduce the vector-based selection strategy, a purely statistical selection technique for commonsense knowledge based on word embeddings.
We compare different commonsense knowledge selection techniques for the purpose of theorem proving and demonstrate the usefulness of vector-based selection with a case study.

\end{abstract}
\section{Introduction}
Current automated theorem provers are high performant and are applied in many areas. In many domains, automated theorem provers have to work on large amounts of background knowledge. This can be large ontologies such as Yago \cite{Suchanek:2008:YLO:1412759.1412998} and SUMO \cite{Pease11}, or mathematical libraries.
 Often the background knowledge is so large that it cannot be considered as a whole by the automated theorem prover.
The observation that even when large sets of background knowledge are used, usually only a very small subset of this knowledge is needed for the proof has led to the development of selection strategies.
The task of selection strategies is: For a large knowledge base and a  problem, determine a (preferably small) subset of the knowledge base for which a proof for the problem can be found.
A selection technique which is broadly used is the SInE \cite{DBLP:conf/cade/HoderV11} selection strategy. The basic idea of SInE is to determine all relevant symbols from the problem and to select axioms based on this relevancy information.
\opt{long}{SInE is successfully used by many provers and applied in many areas.}

In recent years, many sources of commonsense knowledge have emerged such as ontologies like SUMO, and Cyc \cite{lenat1995cyc} and knowledge graphs like ConceptNet \cite{DBLP:conf/aaai/SpeerCH17}. 
\opt{long}{Especially the latter contain large amounts of everyday knowledge and represent an important source of background knowledge for many areas. Even though knowledge graphs cannot be directly processed by automated theorem provers, they can be transformed into first-order logic \cite{DBLP:journals/corr/abs-1912-12957}.}
A nice property of  knowledge bases with commonsense knowledge is that the used symbol names are often based on natural language words. For example in SUMO you find symbol names like \emph{c\_\_SecondarySchool}. Most selection methods like SInE only count the occurrences of symbols. For these methods it is irrelevant whether a symbol is called \emph{c\_\_SecondarySchool} or $p$.

An exception among the selection strategies is Similarity SInE \cite{cade2019}, an extension of SInE which takes the meaning of symbol names into account by using word embeddings. A Word embedding is a mapping from words to vectors in $\mathbb{R}^n$ learned from large text corpora using neural networks. In this process, words that occur in a similar context are mapped to similar vectors. This allows to calculate the similarity of two given words.
Compared to SInE, Similarity SInE does not only consider relevant symbols for the selection but also symbols which are similar to relevant symbols according to a word embedding.
Thus, Similarity SInE is a hybrid selection technique that combines syntax-based selection with statistical methods (word embeddings). 
Experiments in \cite{cade2019} have shown that Similarity SInE is superior to SInE selection in certain domains. A prerequisite for Similarity SInE is that the symbol names used in the knowledge base can be mapped to the vocabulary of the used word embedding.  
For example, the symbol \emph{c\_\_SecondarySchool} mentioned above does not appear in the vocabulary of common word embeddings and must be mapped to \emph{secondary\_school}. 
Quality and coverage of this mapping (the percentage of symbol names mapped to the word embeddings vocabulary) are important for the selection result of Similarity SIne but have not yet been studied in detail. We report on experiments in this regard.
%

Furthermore, we present a new selection strategy, the \emph{vector-based selection} that pays attention to the meaning of symbol names. It is a purely statistical selection in which each axiom of a knowledge base is represented as a vector. This representation makes it is possible to determine the $k$ axioms in the knowledge base most similar to a given proof task. 
\opt{long}{Like in Similarity SInE, a word embedding is used in the vector-based selection strategy. Therefore, a mapping of the symbol names of the knowledge base to the vocabulary of the word embedding is necessary here as well.}
In experiments, we observe that for theorem proving in the commonsense reasoning domain, Similarity SInE leads to the best results and is superior to both SInE and the vector-based selection. 
The main contributions are:
\begin{compactitem}
\item The introduction of the vector-based selection strategy, a purely statistical selection technique for commonsense knowledge.
\item A detailed evaluation comparing different selection techniques for commonsense knowledge for theorem proving purposes.
\item An evaluation of different mappings from symbol names to words in the vocabulary of the used word embedding for Similarity SInE providing insights into desired properties of such mappings.
\item A case study using benchmarks for creativity testing in humans which demonstrates that the vector-based selection allows to select commonsense knowledge in a very focused way.
\end{compactitem}
The paper is structured as follows: after discussing related work in Sec.~\ref{sec:relatedwork} and preliminaries in Sec.~\ref{sec:preliminaries}, we briefly revise the SInE selection strategy in Sec.~\ref{sec:sine}. Next, we turn to the integration of statistical information into selection strategies in Sec.~\ref{sec:statistical_selection} where, after revising distributional semantics and Similarity SInE, we introduce the vector-based selection strategy and the combination of SInE with vector-based selection. In Sec.~\ref{sec:evaluation} we present experimental results. Finally, we discuss future work.

\section{Related Work}\label{sec:relatedwork}
Most strategies for axiom selection are purely syntactic like \cite{DBLP:conf/cade/HoderV11}, \cite{DBLP:journals/japll/MengP09}, and \cite{DBLP:conf/cade/RoedererPS09}. A semantic strategy for axiom selection is \cite{DBLP:conf/cade/SutcliffeP07} which is a model-based approach. This strategy is based on the computation of models for subsets of the axioms and consecutively extends these sets. 
In \cite{DBLP:conf/cade/LiuWWS20} metrics for the evaluation of selection techniques are presented. In addition to that, the paper introduces three new selection strategies based on the relatedness of formulae.
\cite{DBLP:journals/apin/LiuX22} also presents a selection strategy which allows to select the $k$ axioms from a knowledge base most similar to a given problem. For this, different formula metrics are introduced. 
None of the selection methods mentioned so far in this section take the meaning of symbol names into account.

Meaningfulness of symbol names in the semantic web is evaluated in \cite{DBLP:conf/semweb/RooijBBHS16}. The authors come to the conclusion, that the semantics encoded in the names of IRIs (Internationalized Resource Identifiers) carry a kind of social semantics which coincides with the formal meaning of the denoted resource.

Similarity SInE \cite{cade2019} is an extension of SInE selection which uses a word embedding to take similarity of symbols into account. By this mixture of syntactic and statistical methods, Similarity SInE represents a hybrid selection approach. In contrast, the vector-based selection presented in this paper is a purely statistical approach.


With the help of symbol names, \cite{DBLP:conf/nips/Rocktaschel017} defines an extended unification method. The similarities used for this are learned by neural networks. These neural networks allow to determine the proof success with respect to a vector representation of symbol names. It was shown that this method is successful for inductive logic programming.


\cite{DBLP:conf/cade/Tammet19} presents a different but notable approach for dealing with the vast amount of background knowledge in commonsense knowledge bases. \cite{DBLP:conf/cade/Tammet19} does not focus on selection but presents the resolution prover GKC which is optimised for large knowledge bases and was developed as a core system for commonsense reasoning.

\section{Preliminaries and Task Description}\label{sec:preliminaries}
In this paper the reasoning task we focus on assumes a very large set for axioms called knowledge base (KB) to be given. Furthermore a much smaller set of axioms $F_1, \ldots, F_n$  together with a query $Q$ is given. The reasoning task of interest is to show that KB together with the axioms $F_1, \ldots, F_n$ implies the query $Q$.
This corresponds to showing that $F_1\land \ldots \land F_n \rightarrow Q$ is entailed by KB. We stick to the notation of \cite{DBLP:conf/cade/HoderV11} and denote $F_1\land \ldots \land F_n \rightarrow Q$  as \emph{goal}.
We assume that the size of the KB forbids to use the entire KB to show that the goal is entailed by KB using an automated theorem prover (ATP). In order to still be able to show that the goal follows from KB, it is necessary to select the axioms from KB that are needed for this reasoning task. However, identifying these axioms is not trivial, so common selection strategies are based on heuristics and are usually incomplete. This means that it is not always possible to solve the reasoning task with the selected axioms: If too few axioms have been selected, the prover cannot find a proof. If too many were selected, the reasoner may be overwhelmed with the set of axioms and run into a timeout.

In the following we denote the set of all predicate and function symbols occurring in a formula $F$ by $\mathit{sym(F)}$. We slightly exploit notation and use $\mathit{sym(\mathit{KB})}$ for the set of all predicate and function symbols occurring in $\mathit{KB}$.

\section{SInE: a Syntax-Based Selection Strategy}\label{sec:sine}
In \cite{DBLP:conf/cade/HoderV11} the SInE election strategy is introduced which is successfully used by many ATPs. Since this selection strategy does not consider the meaning of symbol names, we classify this strategy as a \emph{syntax-based} selection.
The basic idea of SInE is to determine a set of symbols for each axiom in the KB which is allowed to \emph{trigger} the selection of this axiom. 
For this a \emph{trigger} relation is defined as follows:

\begin{definition}[Trigger relation for the SInE selection\cite{DBLP:conf/cade/HoderV11} ] \label{def:triggers relation}Let KB be a knowledge base, $A$ be an axiom in $\mathit{KB}$ and $s\in \mathit{sym(A)}$ be a symbol. Let furthermore $\mathit{occ(s)}$ denote the number of axioms in which $s$ occurs in KB\opt{long}{ and $t \in \mathbb{R}$, $t \geq 1$}. Then the triggers relation is defined as \\
\begin{equation*}
\mathit{triggers}(s,A) \text{ iff for all symbols } s' \text{ occurring in } A \text{ we have } occ(s) \leq \opt{long}{t \cdot }occ(s')\label{triggers}
\end{equation*}
\end{definition}
Note that an axiom can only be triggered by symbols occurring in the axiom\opt{disable}{ and a symbol $s$ may only trigger an axiom if there is no symbol $s'$ in $A$ that occurs less frequently in KB than $s$. This prevents frequently occurring symbols such as \emph{subClass} and \emph{instanceOf} from being allowed to trigger all axioms they occur in.}\opt{long}{. Parameter $t$ specifies how strict we are in selecting the symbols that are allowed to trigger an axiom. For $t=1$ (the default setting of SInE), a symbol $s$ may only trigger an axiom if there is no symbol $s'$ in $A$ that occurs less frequently in KB than $s$. This prevents frequently occurring symbols such as \emph{subClass} and \emph{instanceOf} from being allowed to trigger all axioms they occur in.}

The \emph{triggers} relation is then used to select axioms for a given goal:
\begin{definition}[Trigger-based selection \cite{DBLP:conf/cade/HoderV11} ]\label{def:trigger-based selection} Let $\mathit{KB}$ be a knowledge base, $A$ be an axiom in $\mathit{KB}$ and  $s \in \mathit{sym(KB)}$. Let furthermore $g$ be a \tchanges{goal to be proven} from $\mathit{KB}$.
\begin{compactenum}
\item If $s$ is a symbol occurring in the goal $g$, then $s$ is 0-step triggered.
\item If $s$ is $n$-step triggered and $s$ triggers $A$ ($\mathit{triggers(s,A)}$), then $A$ is $n + 1$-step triggered. 
\item If $A$ is $n$-step triggered and $s$ occurs in $A$, then $s$ is $n$-step triggered, too.
\end{compactenum}
An axiom or a symbol is called triggered if it is $n$-step triggered for some $n \geq 0$.
\end{definition}
For a given KB, goal $g$  and some $n \in \mathbb{N}$ SInE selects all axioms which are $n$-step triggered. In the following the SInE selection selecting all $n$-step triggered axioms is called SInE with recursion depth $n$.
%

\section{Use of Statistical Information for the Selection of Axioms}\label{sec:statistical_selection}
SInE selection completely ignores the meaning of symbol names. For SInE it makes no difference whether a predicate is called $p$ or $\mathit{dog}$. 
If we consider KBs with commonsense knowledge, the meaning of symbol names provides information that can be exploited by a selection strategy.
For example, the symbol \emph{dog} is more similar to the symbol \emph{puppy} than to the symbol \emph{car}. If a goal containing the symbol \emph{dog} is given, it is more reasonable to select axioms containing the symbol \emph{puppy} than axioms containing the symbol \emph{car}.

\subsection{Distributional Semantics}
To determine the semantic similarity of symbol names, we rely on distributional semantics of natural language, which is used in natural language processing. \opt{long}{The basic idea of distributional semantics is best explained by a quote from Firth, one of the founders of this approach:
 \begin{quote}
You shall know a word by the company it keeps. \cite{Firth}
\end{quote}}
The basis of distributional semantics is the distributional hypothesis \cite{strongContextualHypothesis}, according to which words with similar distributional properties on large texts also have similar meaning.
In other words: Words that occur in a similar context are similar. \looseness=-1

An approach used in many domains which is based on the distributional hypothesis are word embeddings \cite{DBLP:conf/nips/MikolovSCCD13}. Word embeddings map the words of a vocabulary to vectors in $\mathbb{R}^n$.
Typically, word embeddings are learned using neural networks on very large text sets. Since we use existing word embeddings in the following, we do not go into the details of creating word embeddings and refer to \cite{DBLP:journals/corr/abs-1301-3781}.
An interesting property of word embeddings is that semantic similarity of words corresponds to the relative similarities of the vector representations of those words. To determine the similarity of two vector representations the cosine similarity is usually used. 

\begin{definition}[Cosine similarity of two vectors]\label{def:cosine}  Let $u , v \in \mathbb{R}^n$, both non-zero. The \emph{cosine similarity} of $u$ and $v$ is defined as:
\begin{equation*}
\mathit{cos\_sim}(u,v) = \frac{u \cdot v}{\vert\vert u \vert \vert\phantom{\cdot} \vert\vert v\vert \vert} 
\end{equation*}
\end{definition}
\opt{long}{The cosine similarity of two vectors $u$ and $v$ takes values between -1 and 1.}
\opt{long}{For exactly opposite vectors the value is -1, for orthogonal vectors the value is 0 and for equal vectors the value 1.}
The more similar two vectors are, the greater is their cosine similarity.
For example in the ConceptNet Numberbatch Word Embedding \cite{DBLP:conf/aaai/SpeerCH17}, the cosine similarity of \emph{dog} and \emph{puppy} is 0.84140545 which is much larger than the cosine similarity of \emph{dog} and \emph{car} that is 0.13056317. Based on these similarities, word embeddings can be used to determine the $k$ words in the vocabulary most similar to a given word.

\begin{definition}[Set of $k$ vectors most similar to $x_v$ \cite{cade2019}] Let $V$ be a vocabulary and $f$ a word embedding, i.e. ${f:V \rightarrow \mathbb{R}^n}$ , $k \in \mathbb{N}$, $\vert V \vert > k$ and $x_v\in  \mathbb{R}^n$ a vector.  Then $\mathit{simvecs}_f(x_v,k)$, the set of $k$ vectors in $f(V)$ most similar to vector $x_v$, is defined as 

\centerline{
$\simvecs_f(x_v,k)=\left\{
\begin{array}{ll}
\lbrace x_1, \ldots, x_k \rbrace & \text{if } x_v \in f(V)\\
\emptyset& \text{else} 
\end{array}\right.$}

such that 
\begin{compactitem}
\item $\lbrace x_1, \ldots x_k \rbrace \subseteq f(V)$, and $\vert \lbrace x_1, \ldots, x_k \rbrace\vert = k$ and
\item there is no $x_j \in f(V) $ with $x_j \notin \lbrace x_1, \ldots, x_k \rbrace$  and 
$\frac{x_j \cdot x_v}{\vert\vert x_j\vert \vert\phantom{\cdot} \vert\vert x_v\vert \vert} > \frac{x_i \cdot x_v}{\vert\vert x_i\vert \vert\phantom{\cdot} \vert\vert x_v\vert \vert}$
for some $x_i \in \lbrace x_1, \ldots, x_k \rbrace$. 
\end{compactitem}
$\simwords_f(w,k)$, the set of $k$ words most similar to word $w$ in $f$, is defined as
\[\simwords_f(w,k)=\lbrace w'\in V \mid f(w') \in \simvecs_f(f(w),k)\rbrace \]
%
\end{definition}
Word embeddings are used in the selection strategy Similarity SInE \cite{cade2019}, which is briefly revised in the next section.
%

\subsection{Similarity SInE: Enriching a Syntax-Based Selection Strategy with Distributional Semantics}
Similarity SInE \cite{cade2019} is a selection strategy that integrates statistical information on similarities of symbol names into SInE selection. Word embeddings are used as the source for these similarities.
For this, the trigger relation of SInE is extended such that not only the rarest symbol in an axiom is allowed to trigger the axiom but also the $k$ symbols which are most similar to the rarest symbol according to some word embedding.
Assuming that the symbols in KB and the vocabulary of the word embedding coincide, this leads to the following definition:

\begin{definition}[Word embedding enhanced trigger relation \cite{cade2019}] \label{def:simtriggers}
Let $\mathit{KB}$ be a knowledge base, $A$ an axiom in $\mathit{KB}$, $s \in \mathit{sym(KB)}$ be a symbol and $k\in \mathbb{N}$.  Let furthermore $f:V \rightarrow \mathbb{R}^n$ be a word embedding with vocabulary $V = \mathit{sym(KB)}$.
Then the word embedding enhanced set of symbols triggering axiom $A$ is defined as follows: 
\begin{align*}
\simtriggers_f(A,k)  = & \bigcup_{s \in \lbrace s' \mid \mathit{triggers}(s',A) \rbrace} (\lbrace s \rbrace \cup (\simwords_f(s,k)))
\end{align*}
\end{definition}
The trigger-based selection given in Def.~\ref{def:trigger-based selection} remains unchanged.

The assumption, that the symbols of KB and the vocabulary of the word embedding coincide however is not realistic. Therefore, it is necessary to define a mapping of the symbols of KB to the vocabulary of the word embedding (see Sec.~\ref{sec:mapping}).

\subsection{Vector-Based Selection: A Statistical Selection Strategy}
\begin{figure}[t]
    \centering
\begin{tikzpicture}[scale=\textwidth,node distance=4cm,every node/.style={fill=white, font=\sffamily}, align=left]
  \node (KB)           [box]              {\scriptsize{$\mathit{KB}$}};
  \node (WordEmbedding)     [box, right of=KB, xshift=-1cm]          {\scriptsize{word embedding}};
               \node (ProofTask2)     [box, right of=WordEmbedding, xshift=-1cm] {\scriptsize{goal $G$}};
  \node (VectorTransformation)      [activityStarts, below of=KB, yshift=3cm]   {\scriptsize{vector transformation}};
  \node (Selection)      [activityStarts, below of=WordEmbedding,yshift=1.0cm]   {\scriptsize{vector-based selection}};
  \node (VectorTransformationFormula)      [activityStarts, right of=VectorTransformation, xshift=2cm]   {\scriptsize{vector transformation}};
      \node (F3)      [box, below of=Selection, yshift=3cm]   {\scriptsize{$F_3$}};
      \node (F2)      [box, left of=F3, xshift=3cm]   {\scriptsize{$F_2$}};
      \node (F1)      [box, left of=F2, xshift=3cm]   {\scriptsize{$F_1$}};
  \node (F4)      [box, right of=F3, xshift=-3cm]   {\scriptsize{$\ldots$}};
  \node (Fk)      [box, right of=F4, xshift=-3cm]   {\scriptsize{$F_k$}};
        \node (kmost)      [box, below of=F3, yshift=3.7cm]   {\scriptsize{$k$ axioms in $\mathit{KB}$ most similar to $G$}};
    \node (VectorRepresentationQ)     [box, below of=VectorTransformationFormula, yshift=3cm] {\scriptsize{vector representation of $G$}};
    \node (VectorRepresentationKB)     [box, below of=VectorTransformation, yshift=3cm] {\scriptsize{vector representation of $\mathit{KB}$}};
  \draw[->]             (KB) -- (VectorTransformation);
  \draw[->]             (WordEmbedding) -- (VectorTransformation);
    \draw[->]             (WordEmbedding) -- (VectorTransformationFormula);
        \draw[->]             (ProofTask2) -- (VectorTransformationFormula);
    \draw[->]             (Selection) -- (F1);
    \draw[->]             (Selection) -- (F2);
    \draw[->]             (Selection) -- (F3);
    \draw[->]             (Selection) -- (F4);
    \draw[->]             (Selection) -- (Fk);
    \draw[->]              (VectorTransformationFormula) -- (VectorRepresentationQ);
    \draw[->]             (VectorRepresentationQ) -- (Selection);
    \draw[->]              (VectorTransformation) -- (VectorRepresentationKB);
    \draw[->]             (VectorRepresentationKB) -- (Selection);
    \end{tikzpicture}

    \caption{Overview of the vector-based selection strategy. The vector transformation of the KB and the vector transformation of the goal use the same word embedding.}
    \label{fig:vb_selection}
\end{figure}
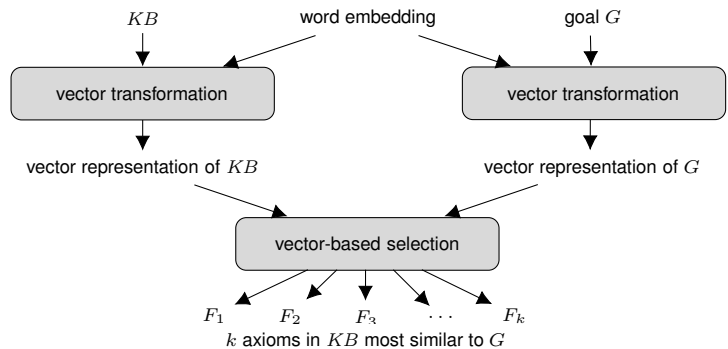

Word embeddings represent words as vectors in such a way that words that are frequently used in a similar context are mapped to similar vectors.
Vector-based selection aims to represent axioms of a KB as vectors in such a way that similar axioms are mapped to similar vectors.
Fig.~\ref{fig:vb_selection} gives an overview of the vector-based selection strategy. In a preprocessing step, vector representations are computed for all axioms of the KB using an existing word embedding. 
This preprocessing step has to be performed only once. 
Given a goal $G$ for which we want to check if it is implied by KB, we transform $G$ into a vector representation using the same word embedding as for the vector transformation of KB.
Next, vector-based selection determines the $k$ vectors in the vector representation of KB most similar to the vector representation of goal $G$. The corresponding $k$ axioms form the result of the selection.
Various metrics can be used for determining the $k$ vectors that are most similar to the vector representation of $G$. We use cosine similarity, which is also widely used in word embeddings.

One way to represent an axiom as a vector is to look up the vectors of all the symbols occurring in the axiom in the word embedding and represent the axiom by the average of these vectors.
However this treats all symbols occurring in an axiom equally. This is not always useful, as the axiom in Fig.~\ref{fig:frequencies} illustrates for which it seems desirable that the symbols \emph{instance}, \emph{agent} and \emph{patient} contribute less to the computation of the vector representation than the symbols \emph{carnivore}, \emph{eating} and \emph{animal}. The reason for this lies in the frequency of the symbols in the KB which are given in the Table in Fig.~\ref{fig:frequencies}. Symbols \emph{carnivore}, \emph{eating} and \emph{animal} occur much less frequently than \emph{instance}, \emph{agent} and \emph{patient} in Adimen SUMO. This suggest, that   \emph{carnivore}, \emph{eating} and \emph{animal}  are more important for the statement of the axiom. This is similar to the idea in SInE that only the least common symbol in an axiom is allowed to trigger the axiom.
%
%
We implement this idea in the computation of the vector representation of axioms by weighting the influence of a symbol using inverse document frequency (idf). In the area of information retrieval, for the task of rating the importance of word $w$ to a document $d$ in a set of documents $D$, idf is often used to diminish the weight of a word that occurs very frequently in the set of documents. Assuming that there is at least one document in $D$, in which $w$ occurs, $\mathit{idf}(w,D)$ is defined as:
\begin{equation*}
\mathit{idf}(w,D) = \log \frac{\vert D \vert}{\vert \lbrace d  \in D \mid w \text{ occurs in } d \rbrace \vert}
\end{equation*}
If $w$ occurs in all documents in $D$, the fraction is equal to 1 and  $\mathit{idf}(w,D)=0$.  If $w$ occurs in only one of the documents in $D$, the fraction is equal to $\vert D \vert$ and $\mathit{idf}(w,D)>0$. The higher the proportion of documents in which $w$ occurs, the lower $\mathit{idf}(w,D)$.

We transfer this idea to KBs by interpreting the KB as a set of documents and each axiom as a document. The resulting computation of idf for a symbol in a KB is given in Def.~\ref{def:idfvectorrep}, where we assume that $\mathit{sym(F)}$ is a subset of the vocabulary of the word embedding. In Sect.~\ref{sec:mapping} we consider the case when this assumption does not hold.
\opt{long}{For the often used tf-idf (term frequency - inverse document frequency) the idf value is multiplied by the term frequency of a term in a certain document. However since the number of occurrences of a symbol in a single axiom does not necessarily correspond to its importance to the axiom (as illustrated by the axiom given in Fig.~\ref{fig:frequencies}),  we omit this multiplication and use idf for the weighting instead.}
\opt{long}{Multiplying the idf value of a symbol with its tf value in a formula could even increase the influence of very frequent symbols like \emph{instance}, since they often appear more than once in a formula. For simplicity, we assume that $\mathit{sym(F)}$ is a subset of the vocabulary of the word embedding. In Sect.~\ref{sec:mapping} we go into more detail about the problem when this assumption does not hold.}

\begin{definition}[idf-based vector representation of a axiom, a KB] \label{def:idfvectorrep} Let $\mathit{KB}$ be a knowledge base with $\mathit{KB}=\lbrace F_1, \ldots, F_n\rbrace$, $n \in \mathbb{N}$, $F \in \mathit{KB}$ be an axiom, 
 $V$ be a vocabulary and $f:V\rightarrow \mathbb{R}^n$ a word embedding. Let furthermore $\mathit{sym(F)} \subseteq V$. 
The \emph{idf} value for a symbol $s \in \mathit{sym}(F)$ w.r.t. $\mathit{KB}$ is defined as
\begin{equation*}
\mathit{idf}(s,\mathit{KB}) = \log \frac{\vert \mathit{KB} \vert}{\vert \lbrace F' \in \mathit{KB} \mid s \in \mathit{sym(F')}\rbrace \vert}
\end{equation*}
The \emph{idf-based vector representation} of $F$ is defined as 
\begin{equation*}
v_\mathit{idf}(F) = \frac{\sum_{s \in \mathit{sym(F)}} \mathit{idf}(s,\mathit{KB}) \cdot f(s)}{\sum_{s \in \mathit{sym(F)}} \mathit{idf}(s,\mathit{KB}) }
\end{equation*}
Furthermore, 
$V_\mathit{idf}(\mathit{KB}) = \lbrace v_\mathit{idf}(F_1), \ldots, v_\mathit{idf}(F_n) \rbrace$ denotes the the idf-based vector representation of $\mathit{KB}$.
\end{definition}
\opt{long}{Note that this definition completely ignores the structure of axioms resulting that axiom $\forall X (\mathit{animal}(X) \land \mathit{fluffy}(X))$ is represented by the same vector as $\forall X (\mathit{anima}l(X) \lor \mathit{fluffy}(X))$. However, this is not a disadvantage, since our goal is a selection of axioms that matches the topic of a goal, and therefore we need the vector representation of an axiom to represent only the topic and not the statement of the axiom.}
\begin{figure}[t]
\centering
\begin{align*}
	\forall  X,Y,Z \Bigl(\bigl(&instance(X,carnivore) \land instance(Y,eating) \\
				&\land agent(Y,X) \land patient(Y,Z) \bigr) \rightarrow instance(Z,animal)\Bigr)
\end{align*}
\small{
\begin{tabular}{lcccccc}
\toprule
\textbf{Symbol Name:}\phantom{some} & instance\phantom{few} & agent \phantom{few}& patient \phantom{few}& carnivore \phantom{few}& eating \phantom{few}& animal\\
 \midrule
\textbf{Frequency:} & 4237\phantom{few} & 140\phantom{few} & 183\phantom{few} & 5\phantom{few} & 6\phantom{few} & 63\\
  \bottomrule
\end{tabular}}
\caption{Example axiom from Adimen SUMO together with frequencies of the symbols of the axiom in Adimen SUMO. To increase readability, we omitted prefixes of symbols.}
\label{fig:frequencies}
\end{figure}

Given a goal $G$ and a KB, we can use the vector representations of KB and $G$ to select the $k$ axioms from KB most similar to the vector representation of $G$ (see Fig.~\ref{fig:vb_selection}).

\begin{definition}[Vector-based selection]\label{def:vb-selection} Let $\mathit{KB}$ be a knowledge base, $G$ be a goal with $\mathit{sym}(G)\subseteq\mathit{sym}(KB)$ and ${f:V \rightarrow \mathbb{R}^n}$ a word embedding. Let furthermore $V_\mathit{KB}$ be a vector representation of $\mathit{KB}$ and $v_G$ a vector representation for $G$ both constructed using $f$. 
For $k \in \mathbb{N}$, $k \leq \vert \mathit{KB} \vert$ the $k$ axioms in KB most similar to $G$ are  given as
\begin{align*}
\mathit{mostsimilar}(\mathit{KB}, G, k) = \lbrace & F_1, \ldots, F_k  \mid \lbrace F_1, \ldots, F_k  \rbrace \subseteq \mathit{KB} \text { and } \\
&\forall F' \in \mathit{KB} \setminus \lbrace F_1, \ldots, F_k \rbrace \\
&\mathit{cos\_sim}(v_{F'},v_G) \leq \min_{i=1,\ldots, k} \mathit{cos\_sim}(v_{F_{i}}, v_G) \rbrace.
\end{align*}
For a knowledge base KB, goal $G$ and some $k \in \mathbb{N}$, vector-based selection selects $\mathit{mostsimilar}(\mathit{KB}, G, k)$.
\end{definition}
Def.~\ref{def:vb-selection} is intentionally very general and allows other vector representations besides idf-based vector representation. Furthermore, the similarity measure $\mathit{cos\_sim}$ can be easily replaced by some other measure like euclidean distance.

\subsubsection{Combining Vector-Based Selection with Syntactic Selection}
When calculating the idf-based vector representation of axioms, very frequent symbols are considered less than rare symbols. If the goal contains rare symbols, it is rather unlikely that axioms containing only very frequent symbols will be selected. However, axioms that contain only frequent symbols can also be very important for finding proofs. An example of such an axiom is: $\forall X,Y,Z ((\mathit{subClass(X,Y)} \land \mathit{subClass(Y,Z)}) \rightarrow \mathit{subClass(X,Z)})$.
To remedy this problem, we suggest the hybrid \vbsine selection strategy.
\begin{definition}[\vbsine selection]\label{def:xyz-selection} Let $\mathit{KB}$ be a knowledge base, $G$ be a goal with $\mathit{sym}(G)\subseteq\mathit{sym(KB)}$ and ${f:V \rightarrow \mathbb{R}^n}$ a word embedding with $\mathit{sym(KB)} \subseteq V$. 
Let $S_{rd}$ be the set of axioms selected by SInE with recursion depth $\mathit{rd}$.
For $\mathit{KB}$, goal $G$, $k \in \mathbb{N}$ and recursion depth $\mathit{rd}$
\emph{\vbsine selection} selects the set of axioms $S_{rd} \cup \mathit{mostsimilar}(\mathit{KB}, G, k)$.
\end{definition}
Both Similarity SInE and \vbsine selection are  hybrid selection strategies. 
Similarity SInE extends the triggers-relation of syntax-based SInE selection strategy with similar symbols.  \vbsine selection uses a vector representation of axioms together with the results of SInE. Therefore, it can be said that \vbsine selection relies even more on statistical information compared to Similarity SInE.




\section{Mapping Symbol Names to the Vocabulary of a Word Embedding}\label{sec:mapping}
In the previous section we assumed the set of symbols in a KB to be a subset of the vocabulary ($V$) of the used word embedding ($f:V \rightarrow \mathbb{R}^n$). However in practice this is not the case and it is necessary to construct a mapping for this. 
Each combination of KB and word embedding requires a specific mapping. As an example we describe how we generated different mappings 
to relate the symbols in Adimen SUMO \cite{lvez2012AdimenSUMORA} to the vocabulary of the ConceptNet Numberbatch word embedding.

Symbols used in Adimen SUMO use prefixes and camelcase for compound terms. For example \emph{secondary school} is represented as \emph{c\_\_SecondarySchool}. In the ConceptNet Numberbatch word embedding compound words are separated by a ``\_''.
To determine a suitable word in the vocabulary of the word embedding for a symbol $s \in \mathit{sym(KB})$ we combined a brute-force step with the Adimen SUMO WordNet mapping. 
The brute-force method constructs $\mathit{bruteForce}(s)$ for a symbol $s$ by removing the above mentioned prefixes, removing some suffixes (like \emph{\_fn} or \emph{\_function}), performs camelcase splitting and converts the result to lower case. 

%

The Adimen SUMO WordNet mapping relates symbols from Adimen SUMO to so called WordNet synsets.
WordNet \cite{DBLP:journals/cacm/Miller95} is a lexical database for the English language where nouns, verbs and adjectives are grouped in synsets (sets of cognitive synonyms). Intuitively, a synset is a set of words with the same or very similar meaning. Fig.~\ref{fig:wordnet} shows some of the noun senses of the word \emph{trunk} in WordNet.
\begin{figure}[t]
\begin{compactitem}
\item S: (n) trunk, tree trunk, bole (the main stem of a tree; usually covered with bark; the bole is usually the part that is commercially useful for lumber)
\item S: (n) trunk (luggage consisting of a large strong case used when traveling or for storage)
\item S: (n) proboscis, trunk (a long flexible snout as of an elephant)
\end{compactitem}
\caption{Some of the noun senses of word \emph{trunk} in WordNet. Each line represents a synset of the word \emph{trunk}. 
The text given in parentheses represents the sense. The synset in the last line contains the synonyms \emph{proboscis} and \emph{trunk} and the sense \emph{a long flexible snout as of an elephant}.}\label{fig:wordnet}
\end{figure}

The WordNet mapping for Adimen SUMO indicates different relationships between symbols and synsets: 
The synonym relationship is the most interesting relationship since it indicates that an Adimen SUMO symbol and a synset and are synonyms. In the following 
$\mathit{wn\_syn}(s)$ denotes the first WordNet synset that symbol $s$ is mapped to in the Adimen SUMO WordNet mapping using the synonym relationship.

Other relationships used in the Adimen SUMO WordNet mapping are the hyponym (sub-concept) and instance relationship. However, these are by their nature much less precise. For example, the Adimen SUMO symbol $c\_\_\mathit{BodyPart}$ is mapped to the last synset for the word \emph{trunk} in Fig.~\ref{fig:wordnet} with the hyponym relationship. This, even though elephant trunks are undoubtedly a part of the body, represents a very imprecise connection. 
\opt{long}{Especially considering that the word \emph{trunk} in ConceptNet Numberbatch Word Embedding combines all synsets for the word \emph{trunk} listed in Fig.~\ref{fig:wordnet}, because Word Embeddings do not distinguish between the different meanings of a word.}
Since the hyponym and the instance relationship are not as precise as the synonym relationship, we give priority to synonym relationship.
In the following $\mathit{wn\_sub}(s)$ ($\mathit{wn\_inst}(s)$) denotes the first synset that symbol $s$ is mapped to in the Adimen SUMO WordNet mapping using the hyponym (instance) relationship.

Tab.~\ref{tab:mappings} gives details on different mappings we considered:
Mapping $M_1$ uses a mapping relation $\mathit{rel}_{M_1}$ defined as:
\begin{align*}
\mathit{rel}_{M_1} = &\lbrace (s,\mathit{bruteForce}(s)) \mid \mathit{bruteForce}(s) \in V\rbrace \cup \\
&\lbrace (s,\mathit{wn\_syn}(s)) \mid \mathit{bruteForce}(s) \notin V\text{ and  }\mathit{wn\_syn}(s) \text{ is defined}\rbrace
\end{align*}
We furthermore performed some semi-automatic improvements of the Adimen SUMO WordNet mapping: First, for symbols without entries to synonyms in the WordNet mapping, possible mapping candidates were determined automatically. For this purpose, lemmatization was performed after prefix and suffix removal and camel case splitting, and suitable synsets were searched for. From the mapping candidates determined in this way, a suitable one was selected manually, if available. This leads to an extension of the above mentioned $\mathit{wn\_syn}$ function which we denote by $\mathit{wn\_syn'}$. 
Mapping $M_2$ presented in Tab.~\ref{tab:mappings} uses the mapping relation $\mathit{rel}_{M_2}$ which corresponds to $\mathit{rel}_{M_1}$ but uses the improved $\mathit{wn\_syn'}$ instead of $\mathit{wn\_syn}$.
This increased coverage (percentage of symbols in $\mathit{sym(KB)}$ for which the mapping is defined) of the mapping from 63.3\% to 76.3\%. 

Mapping $M_3$ in Tab.~\ref{tab:mappings} uses the mapping relation $\mathit{rel}_{M_3}$ which is defined as:
\begin{align*}
\mathit{rel}_{M_3} = & \mathit{rel}_{M_2} \cup \lbrace (s,\mathit{wn\_sub(s)}) \mid  \mathit{rel}_{M_2}(s) \text{ undefined and } \mathit{wn\_sub(s)} \text{ is defined}\rbrace \cup \\
&\lbrace (s,\mathit{wn\_inst}(s)) \mid \mathit{rel}_{M_2}(s), \mathit{wn\_sub(s)} \text{ both undef. and } \mathit{wn\_inst(s)} \text{ is defined}\rbrace
\end{align*}
\begin{table}[t]
\centering
\small{
\begin{tabular}{lcl}
\toprule
 & \textbf{Coverage} & \textbf{Construction}\\
 \midrule
$M_1$ & 63.3\% &Brute-Force + WordNet mapping synonyms\\
$M_2$ & 76.3\%&Brute-Force + improved WordNet mapping synonyms\\
$M_3$ & 81.6\%&Brute-Force +  improved WordNet mapping synonyms + WordNet mapping\\
& &  hyponyms and instances\\
  \bottomrule
\end{tabular}}
\caption{Overview of the different mappings considered.}
\label{tab:mappings}
\end{table}
Mappings can be easily integrated into selection strategies. See \cite{cade2019} for details how to do this for Similarity SInE.
To integrate a mapping $\mathit{rel}$ in the computation of the vector representation of axiom $F$ in vector-based selection, we have to replace the formula for $v_\mathit{idf}(F)$ in Def.~\ref{def:idfvectorrep} by 
$v_\mathit{idf}(F) = \frac{\sum_{s \in \mathit{mapsym(F)}} \mathit{idf}(s,\mathit{KB}) \cdot f(\mathit{rel}(s))}{\sum_{s \in \mathit{mapsym(F)}} \mathit{idf}(s,\mathit{KB}) }$ with $\mathit{mapsym}(F)$ the symbols in $\mathit{F}$ for which $\mathit{rel}$ is defined.

Obviously, the result of Similarity SInE selection strongly depends on the used mapping. If the mapping has a low coverage only few similarities can be used. In the worst-case, if the mapping is empty, Similarity SInE behaves like the SInE. 
For  vector-based selection, the mapping is even more important. Every symbol which is not mapped to \opt{long}{the vocabulary $V$ of the used word embedding}\opt{disable}{$V$} is ignored in the calculation of the vector representation. In the worst case, if no symbol in an axiom can be mapped to $V$, it is not possible to represent the axiom as a vector and the axiom will never be selected by the vector-based selection.

Considering Adimen SUMO (7,432 axioms in total): For mapping $M_2$ for 49 and for mapping $M_3$ for 36 axioms it was not possible to compute a vector representation since none of their symbols could be mapped to the vocabulary of the word embedding. 
Since mapping $M_3$ additionally draws on information about hyponym and instances from Adimen SUMO's WordNet mapping, $M_3$ naturally has higher coverage than $M_2$. However, it is to be expected that the pairs added to $M_3$ compared to $M_2$ are rather of inferior quality.
\opt{long}{In Sec.~\ref{sec:evaluation}, we will examine the effect of the mapping relations presented in Tab.~\ref{tab:mappings} with Similarity SInE, vector-based selection, and \vbsine selection.}

%

%
%

\section{Evaluation}\label{sec:evaluation}
To evaluate the different selection strategies, we need benchmark problems where one fixed KB is used for a number of goals. Furthermore, the KB has to be sufficiently large such that selection is necessary. 
 Similarity SInE as well as the vector-based selection strategy heavily rely on word embeddings. In order to evaluate these selection strategies, it is necessary that the symbols used in the KB can be mapped to the vocabulary of the used word embedding. This could be achieved, for example, through an existing WordNet mapping. Unfortunately, there is no WordNet mapping for the problems in the TPTP, making it difficult to use the TPTP problems for evaluation. 
\opt{long}{Generating such a mapping automatically for problems in the TPTP is difficult because the problems in the TPTP come from different sources and therefore there is no uniform naming of predicate and function symbols.}

For our experiments, we choose Adimen SUMO which is a first-order logic translation of the SUMO ontology. Adimen SUMO lends itself to our experiments for several reasons: 
It comes with 8,010 automatically generated white-box truth tests, which are supposed to be entailed by the KB. Since Adimen SUMO is very large, selection is necessary for many proof tasks. The WordNet mapping that is available for SUMO can also be used for Adimen SUMO.

The selection of axioms is completely independent from the reasoner used afterwards. This allows us to use different reasoners for our experiments. We choose E \cite{DBLP:conf/cade/KaliszykSUV15}, a highly optimised saturation-based first-order prover and Hyper \cite{cadesd}, a tableau prover which has a compact proof structure and efficient equality handling.

For the evaluation, we first determined the set of Adimen SUMO white-box tests for which Hyper was unable to find a proof within 15 seconds without selection. From this set of problems, 1,000 problems were randomly selected for evaluation.
Analogously, such a set of 1,000 problems was determined for E.
Note that the 1,000 problems for the evaluation of Hyper is not the same as the set of 1,000 problems for the evaluation of E.
Since Hyper was able to find proofs for a much smaller fraction of the Adimen SUMO whitebox tests without selection than E, it can be assumed that the problems for which E was unable to find proofs are more difficult problems than the problems for which Hyper was unable to find a proof. In the following evaluation, however, we are not interested in comparing the provers, but in comparing the different selection strategies. The goal is to investigate how the selection strategies presented above affect the number of proofs found by a prover.

All experiments presented below were carried out on a computer featuring an Intel(R) Xeon(R) CPU E5-2680 v3 @ 2.50GHz, 2000 MHz (only two cores were used) with 48 GB RAM. For both provers, we used a timeout of 15 seconds (cpu time). This timeout is based on the experimental results from \cite{DBLP:journals/corr/AlvezHLR17}, where for the majority of the Adimen-SUMO problem's proofs were found in up to 10 seconds.


\subsection{Different Mappings in Similarity SInE}\label{sec:eval_mappings}
In Sec.~\ref{sec:mapping} we introduced three ways to map symbols of Adimen SUMO to the vocabulary of the used word embedding. 
In general, mappings differ mainly in the coverage of the mapped symbols as well as in the quality of the mapping. 
Of course, a mapping with high coverage and high quality would be best. In practice, however, it is often not possible to automatically increase the coverage arbitrarily without decrease in quality.
\begin{figure}[t]
\begin{subfigure}{.5\textwidth}
\begin{tikzpicture}[scale=0.7]
\begin{axis} [xticklabels={proof, model, timeout},xtick={1, 2, 3}, ybar, enlarge x limits=0.25, ymin=0,ymax=1000,ylabel= number of problems Hyper gave the specified result, legend style={at={(0.5,-0.2)},anchor=north}, nodes near coords, point meta=y, every node near coord/.append style={font=\tiny}]
 \addplot coordinates {
    (1,13)
    (2,589)
    (3,398)
    };
    \addplot coordinates {
    (1,399)
    (2,047)
    (3,554)
    };
    \addplot coordinates {
    (1,538)
    (2,028)
    (3,434)
    };
    \addplot coordinates {
    (1,163)
    (2,024)
    (3,813)
    };
\legend{SInE,Similarity SInE Selection with mapping $M_1$,Similarity SInE Selection with mapping $M_2$, Similarity SInE Selection with mapping $M_3$}  
\end{axis}
\end{tikzpicture}
\end{subfigure}
\begin{subfigure}{.5\textwidth}
\begin{tikzpicture}[scale=0.7]
\begin{axis} [xticklabels={proof, model, timeout},xtick={1, 2, 3},  enlarge x limits=0.25,ybar,ymin=0,ymax=1000,ylabel=  number of problems E gave the specified result, legend style={at={(0.5,-0.2)},anchor=north},nodes near coords, point meta=y, every node near coord/.append style={font=\tiny}]
    \addplot coordinates {
    (1,14)
    (2,849)
    (3,140)
    };
    \addplot coordinates {
    (1,084)
    (2,661)
    (3,255)
    };
    \addplot coordinates {
    (1,265)
    (2,466)
    (3,269)
    };
    \addplot coordinates {
    (1,271)
    (2,454)
    (3,275)
    };
\legend{SInE,Similarity SInE Selection with mapping $M_1$,Similarity SInE Selection with mapping $M_2$, Similarity SInE Selection with mapping $M_3$}  
\end{axis}
\end{tikzpicture}
\end{subfigure}
\caption{Results of Hyper and E on 1000 randomly selected Adimen SUMO problems that Hyper or E could not solve without selection (see Sec.~\ref{sec:evaluation}). For these problems, SInE as well as Similarity SInE (with $k=100$) with three different mappings ($M_1, M_2$ and $M_3$) was used. For all selections, recursion depth 1-6 was used. \opt{long}{Note that Hyper and E worked on different sets of 1,000 problems, since they could also solve different sets of problems without selection (see Sec.~\ref{sec:evaluation}).}
}
  \label{fig:mapping}
\end{figure}
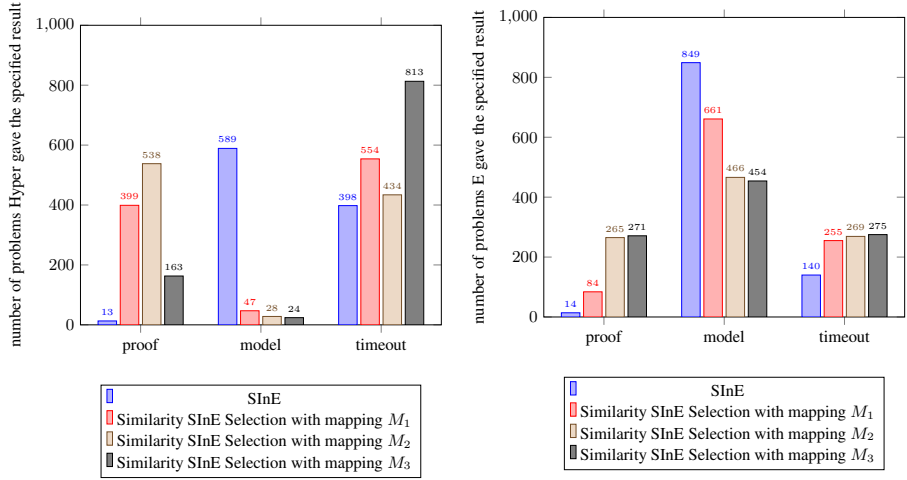
To evaluate the influence of coverage and quality of mappings on Similarity SInE, we conducted experiments on the 1,000 randomly determined Adimen SUMO problems described above. We used Similarity SInE with mappings $M_1, M_2$ and $M_3$ presented in Tab.~\ref{tab:mappings} to select for these problems and ran E or Hyper on the result of the selection.  
Fig.~\ref{fig:mapping} presents the results of these experiments. For comparison, Fig.~\ref{fig:mapping} also shows the results of a run with pure syntactic selection (SInE). 
Regardless of the prover, Similarity SInE is superior to SInE.

As described in Sec.~\ref{sec:mapping}, the difference between $M_1$ and $M_2$ is in coverage: mapping $M_2$ is an extension of $M_1$ with high-quality semi-automatically generated information. 
\opt{long}{Therefore, it is not surprising that Hyper and E find more proofs on the problems selected with Similarity SInE with $M_2$ than on the problem selected with Similarity SInE with $M_1$.}
\opt{disable}{We observe that E finds more than three times as much proofs using $M_2$ in Similarity SInE compared to using $M_1$. Hyper also finds more proofs using $M_2$ in Similarity SInE compared to using $M_1$ although the increase here is not quite as high. We conclude that increasing coverage of the mapping while preserving the quality improves the results  of Similarity SInE and thus leads to more found proofs.}
\opt{long}{E finds 84 proofs when using $M_1$ in Similarity SInE and 265 proofs when using $M_2$. This more than triples the number of proofs found.
Compared to this, Hyper finds 399 proofs when using $M_1$ in Similarity SInE and 538 proofs when using $M_2$. This corresponds to an improvement by a factor of 1.3. We conclude that increasing coverage of the mapping while preserving the quality improves the results  of Similarity SInE selection and thus leads to more proofs by provers.}

Mapping $M_3$ represents an extension of $M_2$ with less precise information about hyponyms and instances from the Adimen SUMO WordNet mapping. 
Hyper finds 538 proofs when using $M_2$ with Similarity SInE selection and only 163 proofs when using $M_3$ with Similarity SInE. The number of timeouts increases from 434 with $M_2$ to 813 with $M_3$.
Compared to that, the use of the less accurate mapping $M_3$ does not cause E to find less proofs. 
However, the number of proofs found increases only very slightly (by 6). Also, the number of timeouts encountered in E increases by 6 by using $M_3$ (compared to $M_2$).

In summary, it can be said that increasing the coverage of the mapping used only pays off if a lot of attention is paid to the quality of the mapping and only high-quality additions are made.
%
\opt{long}{
\subsection{Similar Symbols in Similarity-SInE:  \emph{The more the bettter}?}
In the next experiments, we investigate the effect of the number of similar symbols used in the Similarity SInE selection (parameter $k$). 
Please note that for Similarity SInE, increasing the value of $k$ can never result in selecting less axioms. On the contrary, usually increasing $k$ results in the selection of more axioms.

We build on the results of the evaluation of the different mappings in the previous section and now use mapping $M_2$ for our experiments. 
In the previous section, for E the difference between $M_2$ and $M_3$ turned out to be only very small. Therefore, we performed the experiments on similar symbols for E using mapping $M_3$ as well. The results differ only slightly from the ones using $M_2$ and can be found in Fig.~\ref{fig:E_M3} in Appendix~\ref{app:experiments}.

\begin{figure}[t]
\begin{subfigure}{.5\textwidth}
\begin{tikzpicture}[scale=0.7]
\begin{axis} [xticklabels={proof, model, timeout},xtick={1, 2, 3}, ybar, enlarge x limits=0.25, ymin=0,ymax=1000,ylabel= number of problems Hyper gave the specified result, legend style={at={(0.5,-0.2)},anchor=north}, nodes near coords, point meta=y, every node near coord/.append style={font=\tiny}]
 \addplot coordinates {
    (1,257)
    (2,48)
    (3,695)
    };
    \addplot coordinates {
    (1,538)
    (2,28)
    (3,434)
    };
    \addplot coordinates {
    (1,436)
    (2,20)
    (3,544)
    };
\legend{Similarity SInE Selection with mapping $M_2$ and $k=50$\phantom{0},Similarity SInE Selection with mapping $M_2$ and $k=100$, Similarity SInE Selection with mapping $M_2$ and $k=150$}  
\end{axis}
\end{tikzpicture}
\end{subfigure}
\begin{subfigure}{.5\textwidth}
\begin{tikzpicture}[scale=0.7]
\begin{axis} [xticklabels={proof, model, timeout},xtick={1, 2, 3}, ybar, enlarge x limits=0.25, ymin=0,ymax=1000,ylabel= number of problems E gave the specified result, legend style={at={(0.5,-0.2)},anchor=north}, nodes near coords, point meta=y]
 \addplot coordinates {
    (1,63)
    (2,709)
    (3,228)
    };
    \addplot coordinates {
    (1,265)
    (2,466)
    (3,269)
    };
    \addplot coordinates {
    (1,276)
    (2,445)
    (3,272)
    };
\legend{Similarity SInE Selection with mapping $M_2$ and $k=50$\phantom{0},Similarity SInE Selection with mapping $M_2$ and $k=100$, Similarity SInE Selection with mapping $M_2$ and $k=150$}
\end{axis}
\end{tikzpicture}
\end{subfigure}
\caption{Results of Hyper and E on 1,000 randomly selected Adimen SUMO problems that the respective prover could not solve without selection. For these problems, Similarity SInE using mapping $M_2$ with different values for parameter $k$ was used. The left figure shows the results for Hyper, the right figure the results for E.
}
  \label{fig:k}
\end{figure}
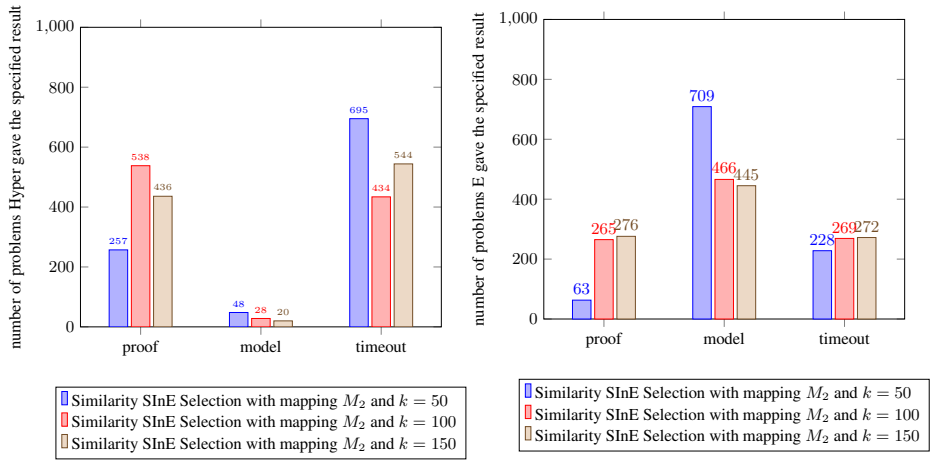

The left side of Fig.~\ref{fig:k} shows the results Hyper achieved on the 1,000 randomly selected Adimen SUMO problems. As a selection technique, Similarity SInE with mapping $M_2$ with different values for parameter $k$ was used. 
We observe that Hyper is able to find only 257 proofs using $k=50$. For $k=100$, the number of proofs found is 538, more than twice as high as for $k=50$. This can also be observed in the experiments with E shown on the right side of Fig.\ref{fig:k}. E is able to find 63 proofs using $k=50$ and for $k=100$ the number of proofs found increases to 265 which is more than a fourfold increase. 
Further increasing the value of $k$ to 150 has different effects for the two provers: For Hyper it results in the much lower number of proofs of 436. At the same time, the number of timeouts for Hyper increases which could be explained by an overloading of the prover by irrelevant axioms in the selection output.
In contrast to that, the number of proofs E is able to find after selecting with $k=150$ slightly increases to 276.

This difference in the two provers can probably be explained by the fact that E is significantly more performant and therefore not overwhelmed by a large set of selected axioms as quickly as Hyper.  This has already shown before, when E was able to solve a much larger set of Adimen SUMO whitebox tests without selection than Hyper. 
But even E no longer shows a large increase in the number of proofs found, illustrating that the results cannot be arbitrarily improved by increasing the $k$ parameter.
This intuitively makes sense, since the $k$ parameter determines the number of similar symbols added to the trigger relation of Similarity SInE. For this a word embedding is used to determine the  $k$ words most similar to the symbol.  Above a certain value for $k$, there is simply not enough similarity between the words in a word embedding.
}

%

\subsection{Vector-Bases Selection}
Next we evaluate vector-based selection and vb-union-SInE selection. For the experiments, we used both Hyper and E on the set of 1,000 randomly selected Adimen SUMO problems described at the beginning of Sec.~\ref{sec:evaluation}.

As a reminder, vb-union-SInE selection involves performing two independent selections (one with vector-based selection, one with SInE), constructing the union of the results of these two selections, and then passing this union to the prover.
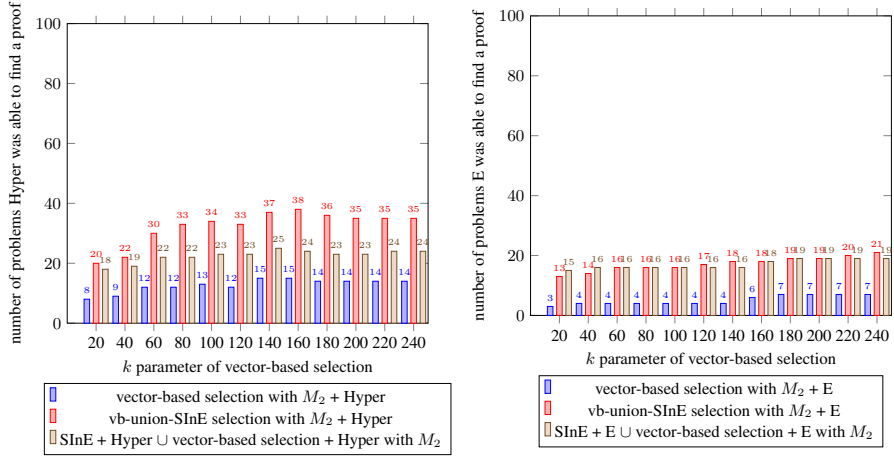
\begin{figure}[t]
\begin{subfigure}{.5\textwidth}
\begin{tikzpicture}[scale=0.7]
\begin{axis} [xtick={20,40,60,80,100,120,140,160,180,200,220,240},xticklabels={20,40,60,80,100,120,140,160,180,200,220,240},ybar, xmin=20,	bar width=3pt,
ymin=0,ymax=100, xmin=0,xmax=250,xlabel = $k$ parameter of vector-based selection, ylabel=  number of problems Hyper was able to find a proof, legend style={at={(0.5,-0.2)},anchor=north},nodes near coords, point meta=y, every node near coord/.append style={font=\tiny},legend style={cells={align=left}}]
    \addplot table [x=a, y=vb, col sep=comma] {data_hyper_m2.csv};
    \addplot table [x=a, y=sinevb, col sep=comma] {data_hyper_m2.csv};
    \addplot table [x=a, y=vereinigung, col sep=comma] {data_hyper_m2.csv};
    \legend{vector-based selection with $M_2$ + Hyper, \vbsine selection with $M_2$ + Hyper, SInE + Hyper $\cup$ vector-based selection + Hyper with $M_2$}  
\end{axis}
\end{tikzpicture}
\end{subfigure}
\begin{subfigure}{.5\textwidth}
\begin{tikzpicture}[scale=0.7]
\begin{axis} [xtick={20,40,60,80,100,120,140,160,180,200,220,240},xticklabels={20,40,60,80,100,120,140,160,180,200,220,240},ybar, xmin=20,	bar width=3pt,
ymin=0,ymax=100, xmin=0,xmax=250,xlabel = $k$ parameter of vector-based selection, ylabel=  number of problems E was able to find a proof, legend style={at={(0.5,-0.2)},anchor=north},nodes near coords, point meta=y, every node near coord/.append style={font=\tiny},legend style={cells={align=left}}]
    \addplot table [x=a, y=vb, col sep=comma] {data_eprover_m2.csv};
    \addplot table [x=a, y=sinevb, col sep=comma] {data_eprover_m2.csv};
    \addplot table [x=a, y=vereinigung, col sep=comma] {data_eprover_m2.csv};
    \legend{vector-based selection with $M_2$ + E, \vbsine selection with $M_2$ + E, SInE + E $\cup$ vector-based selection + E with $M_2$}  
\end{axis}
\end{tikzpicture}
\end{subfigure}
\caption[]{Results of Hyper and E on 1,000 randomly selected Adimen SUMO problems that the respective prover could not solve without selection. The experiments consider mapping $M_2$ and three different selections followed by runs of the two provers: \emph{vector-based selection}, \emph{\vbsine} and the afore described \emph{SInE + prover} $\cup$ \emph{vector-based selection + prover}.
For each setting, we provide the number of proofs found by the respective prover. 
Please note: To increase readability, we have set the maximum value of the y-axis to 100 which should be taken into account when comparing with diagrams from the previous section.
}
  \label{fig:vb_mapping}
\end{figure}

For comparison, we also included experiments in which we performed two completely independent selections followed by two independent runs of a prover: one selection were made with SInE and the prover was called on the result. The second selection was done with vector-based selection and the prover was called on the result. The number of problems for which at least one of the prover runs found a proof was then counted. In Fig.~\ref{fig:vb_mapping}, this is denoted as \emph{SInE + prover} $\cup$ \emph{vector-based selection + prover}, where prover is replaced with the prover used in the depicted experiment.

Comparing the results of calling a prover on the output of vb-union-SInE selection and \emph{SInE + prover} $\cup$ \emph{vector-based selection + prover}  allows us to see whether the combination of statistical and syntactical selection as done in vb-union-SInE is indeed useful or whether possible improvements in the results can be explained by mere addition effects.\looseness=-1

Recall that for the problems considered for Hyper in Fig.~\ref{fig:vb_mapping}, Hyper was able to find 13 proofs after selecting with SInE (see Fig.~\ref{fig:mapping}). Comparing this with the results for Hyper on the left side of Fig.~\ref{fig:vb_mapping}, we see that Hyper performs comparable after selecting with SInE or vector-based selection for values of $k > 60$. 
In comparison, Hyper finds significantly more proofs after selection with \vbsine selection than the 13 after selection with SInE.  This is true 
for all considered values of $k$. Comparing these results with \emph{SInE + Hyper} $\cup$ \emph{vector-based selection + Hyper}, shows that the improvement by \vbsine selection goes beyond mere addition effects.
However, the number of proofs found falls far short of the 538 proofs found using Similarity SInE with mapping $M_2$ (see Fig.~{\ref{fig:mapping}).

E was able to find proofs for 14 of the problems after selection with SInE (see Fig.~\ref{fig:mapping}). Comparing this number with the results of E on the output of vector-based selection (see the right side of Fig.~\ref{fig:vb_mapping})  shows that vector-based selection is inferior to SInE 
for all values of $k$ considered in the experiments. Furthermore, we see that when using E as a prover, \vbsine selection is superior to vector-based selection. This is similar to the results of the evaluation using Hyper.
However, the better results for \vbsine selection can be completely explained by addition effects, which can be seen by comparing with \emph{SInE + E} $\cup$ \emph{vector-based selection + E}. Furthermore, comparing the results to the results of Similarity SInE given in Fig.~{\ref{fig:mapping} shows that selecting with Similarity SInE leads to far more proofs.

The results presented in Fig.~\ref{fig:vb_mapping} use mapping $M_2$. Performing the same experiments using mapping $M_3$ leads to similar results (see Fig.~\ref{fig:vb_mapping_M3} in Appendix~\ref{sec:vb_M3}).

\subsection{Case Study on Commonsense Knowledge} 
In many areas where commonsense knowledge is used as background knowledge, ATPs are used not only for finding proofs, but also as inference engines.
One reason for this is that even if there are large ontologies and knowledge bases with commonsense knowledge, this knowledge is still incomplete.
Nevertheless, ATPs can be very helpful on commonsense knowledge, because the inferences that a prover can draw from a problem description and selected background knowledge provide valuable information.
How well these inferences fit the problem description depends strongly on the selected background knowledge. Here it is also very important that the selected background knowledge is broad enough but still focused.
On example where this is the case is the CoRg project \cite{SSS19a} which aims at modelling the human ability to draw meaningful conclusions in everyday situations with the help of large amounts of background knowledge, ATPs and machine learning.

To evaluate whether the presented selection strategies are appropriate for commonsense knowledge, we use the functional Remote Association Tasks (fRAT) \cite{fRAT} which were developed to measure human creativity. 
In fRAT, three words like \emph{tulip}, \emph{daisy} and \emph{vase} are given and the task is to find a fourth connecting word, called target word (here \emph{flower}). 
The words are chosen in such a way that a functional connection must be found between the three words and the target word.
The solution of the above fRAT task requires the background knowledge that tulips and daisies are flowers and that a vase is a container in which flowers are kept. 

The dataset used for this evaluation was introduced in \cite{DBLP:journals/kbs/OlteteanuSS19} and consists of 48 fRAT tasks. The left side of Fig.~\ref{fig:frat_ex+graph} gives some examples for tasks in the dataset.
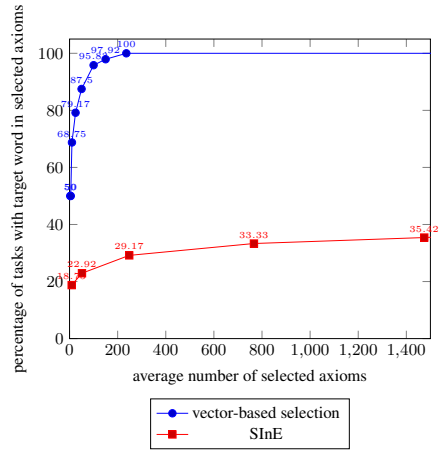
\begin{figure}[t]
\begin{subfigure}{.5\textwidth}
\begin{tabular}{cc} 
\toprule
\thead{Query Words\\$w_1$, $w_2$ and $w_3$} & \thead{Target Word\\$w_t$}\\ 
\midrule 
tulip, daisy, vase &  flower\\
sensitive, sob, weep& cry\\
algebra, calculus, trigonometry &  math\\
duck, sardine, sinker &  swim\\
finger, glove, palm & hand\\
\bottomrule
\end{tabular}
\end{subfigure}
\begin{subfigure}{.5\textwidth}
\begin{tikzpicture}[scale=0.7]
\begin{axis} [xmin=0, xmax=1500,
ymin=0,ymax=105,xlabel = average number of selected axioms, ylabel=  percentage of tasks with target word in selected axioms, legend style={at={(0.5,-0.2)},anchor=north},nodes near coords, point meta=y, every node near coord/.append style={font=\tiny},legend style={cells={align=left}}]
    \addplot table [x=k, y=percent, col sep=comma] {data_frat_vb.csv};
        \addplot table [x=numaxioms, y=percent, col sep=comma] {data_frat_sine.csv};
    \legend{vector-based selection,SInE}  
\end{axis}
\end{tikzpicture}
\end{subfigure}
\caption{On the left: Examples from the fRAT dataset. Given the three query words, the task is to determine the target word which establishes a functional connection. On the right: Percentage of fRat tasks for which the target word occurs in the selected axioms depending on the number of selected axioms. Vector-based selection and SInE were used for the selection.}\label{fig:frat_ex+graph}
\end{figure}
For an fRAT task consisting of the words $w_1$, $w_2$, $w_3$ and the target word $w_t$, we first generate a simple goal $\forall X (w_1(X) \land w_2(X) \land w_3(X))$ using the query words of the tasks as predicate symbols and then select for this goal using different selection strategies. Then we check whether the word $w_t$ occurs in the selected axioms. 
Since we only want to evaluate selection strategies on commonsense knowledge, we do not use a reasoner in the following experiments.

As background knowledge we use ConceptNet \cite{DBLP:conf/aaai/SpeerCH17} which is a knowledge graph containing broad commonsense knowledge in the form of triples. 
For this evaluation, we use a first-order logic translation of around 125,000
of the English triples of ConceptNet as KB (see \cite{DBLP:journals/corr/abs-1912-12957} for details on the translation). 

We use both vector-based selection and SInE to select axioms for the goals created for the fRAT tasks and then check if the target word $w_t$ occurs in the selected axioms. The left side of Tab.~\ref{tab:frat} shows the results for vector-based selection, the right side for SInE. 
Note that for vector-based selection the $k$ parameter naturally determines the number of axioms contained in the result of the selection. 
Since the selected axioms are sorted in descending order with respect to the similarity to the goal in vector-based selection, the left side of Tab.~\ref{tab:frat} furthermore provides the average position of the target word in the selected axioms. 

\begin{table}[t]
\small{
\begin{tabular}{cc}
    \begin{minipage}{.5\linewidth}
   \begin{tabular}{lcc} 
\toprule
\multicolumn{3}{l}{\textbf{Vector-based selection on fRAT}}\\  
\midrule 
\textbf{k} & \phantom{di}\thead{\% of tasks with\\$w_t$ in selection} & \phantom{di}\thead{avg. pos. of\\ target word} \\ 
\midrule 
5 & 50\% & 1.63 \\
10 & 68.75\% & 2.70 \\
25 & 79.17\% & 4.5 \\
50 & 87.5\% & 5.85  \\
100 & 95.83\% & 11.15 \\
$\geq$235 & 100\% & 17.63\\
\bottomrule
\end{tabular}
    \end{minipage} &

    \begin{minipage}{.5\linewidth}
        \begin{tabular}{lcc} 
\toprule
\multicolumn{3}{l}{\textbf{SInE on fRAT}}\\  
\midrule 
\thead{rec.\\depth} & \phantom{di}\thead{\% of tasks with\\$w_t$ in selection} & \phantom{di}\thead{avg. number of\\ selected axioms}\\ 
\midrule 
1 & 18.75\% & 8.92  \\
2 & 22.92\% & 51.69  \\
3 &29.17\% & 248.10 \\
4 & 33.33\% & 766.52 \\
5 & 25.42\% & 1474.00 \\
6 & 37.5\% & 2045.88 \\
\bottomrule
\end{tabular}
    \end{minipage} 
\end{tabular}
\caption{Results of selecting for the 48 fRAT tasks: percentage of tasks where the target word $w_t$ occurs in the selected axioms. Vector-based selection on the left, SInE on the right.}\label{tab:frat}
}
\end{table}
%
%
The results for SInE on the right side of Tab.~\ref{tab:frat} show that  even for recursion depth 6, were SInE selected 2,046 axioms on average for an fRAT task, in only 37.5\% of the tasks the target word occurred in the selection. Compared to that, the result of vector-based selection of only three axioms already contains the target word in 50\% of the tasks. As soon as the vector-based selection selects more than 235 axioms, the target word is contained in the selection for all of the tasks.
The right side of Fig.~\ref{fig:frat_ex+graph} illustrates the relationship between the number of axioms selected and the percentage of target words found for the two selection strategies.
Although SInE selects significantly more axioms than vector-based selection, axioms containing the target word are often not selected. In contrast, vector-based selection is much more focused and even small sets of selected axioms contain axioms mentioning the target word. 


The experiments reveal another problem specific for the task of selecting background knowledge from commonsense KBs: Since KBs in this area usually get extremely large, it is reasonable to assume that a user looking for background knowledge for a set of keywords is not aware of the exact symbol names used in the KB. 
\opt{long}{Therefore it can easily happen that a user looks for background knowledge for a set of words which do not coincide with the symbol names used in the KB.} For example none of the query words \emph{tulip}, \emph{daisy} and \emph{vase} corresponds to a symbol name in our first-order logic translation of ConceptNet. Therefore a selection using SInE with the goal created from these query words results in an empty selection.
In contrast to that, vector-based selection constructs a query vector from the symbol names occurring in the goal (idf-based selection can assume the average idf value for unknown symbols) and selects the $k$ most similar axioms even though the query words from the fRAT task do occur as symbol names in KB. As long as the query words occur in the vocabulary of the used word embedding or can be mapped to this vocabulary, it is possible to construct the query vector and select axioms. 

Experiments in this area using Similarity SIne will be considered in future work. 
We expect Similarity SInE to perform better than SInE w.r.t. finding the target word in the result of the selection. However we do not expect Similarity SInE to be as focused as vector-based selection since Similarity SInE naturally always selects a superset of the selection SInE would perform. \looseness=-1

%
\section{Conclusion and Future Work} 
In KBs representing commonsense knowledge, symbol names usually have a meaning which can be exploited by selection strategies.
In this paper, we presented and compared different selection strategies for commonsense knowledge. All of these selection strategies integrate word embeddings into the selection process.
We presented the vector-based selection strategy, a purely statistical selection strategy together with \vbsine selection, which combines vector-based selection with SInE. 
While our experiments show that the vector-based selection is inferior to Similarity SInE \cite{cade2019}  when it comes to selection for proof tasks, a case study on benchmarks for testing human creativity shows that vector-based selection is able to select very focused on commonsense knowledge. 
In future work, we want to evaluate the usefulness of the introduced selection strategies using benchmarks from the commonsense reasoning area like COPA \cite{DBLP:conf/aaaiss/RoemmeleBG11}.\looseness=-1

\bibliographystyle{abbrv}
\bibliography{referenz}

\appendix
\section{Additional Experimental Results}\label{app:experiments}
\subsection{Similar Symbols Similarity SInE: Experiments with E and Mapping $M_3$}
\begin{figure}[H]
\begin{center}
\begin{tikzpicture}[scale=0.7]
\begin{axis} [xticklabels={proof, model, timeout},xtick={1, 2, 3},  enlarge x limits=0.25,ybar,ymin=0,ymax=1000,ylabel=  number of problems E gave the specified result, legend style={at={(0.5,-0.2)},anchor=north}, nodes near coords, point meta=y  
]
    \addplot coordinates {
    (1,55)
    (2,708)
    (3,237)
    };
    \addplot coordinates {
    (1,271)
    (2,454)
    (3,275)
    };
    \addplot coordinates {
    (1,281)
    (2,433)
    (3,286)
    };
\legend{Similarity SInE Selection with mapping $M_3$ and $k=50$\phantom{0},Similarity SInE Selection with mapping $M_3$ and $k=100$, Similarity SInE Selection with mapping $M_3$ and $k=150$}
\end{axis}
\end{tikzpicture}
\caption{Results of E on 1,000 randomly selected Adimen SUMO problems that E could not solve without selection. For these problems, Similarity SInE with different values for parameter $k$ and mapping $M_3$ was used. 
}
  \label{fig:E_M3}
  \end{center}
\end{figure}
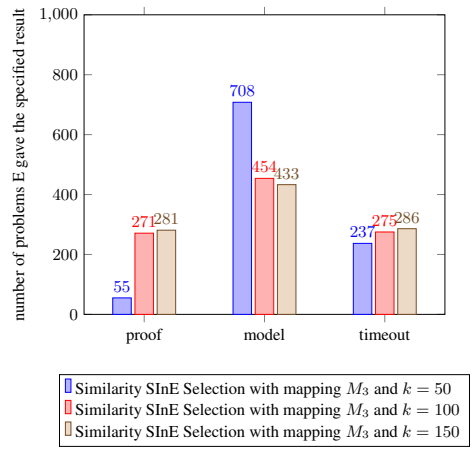

\subsection{Vector-Based Selection with Mapping $M_3$}\label{sec:vb_M3}
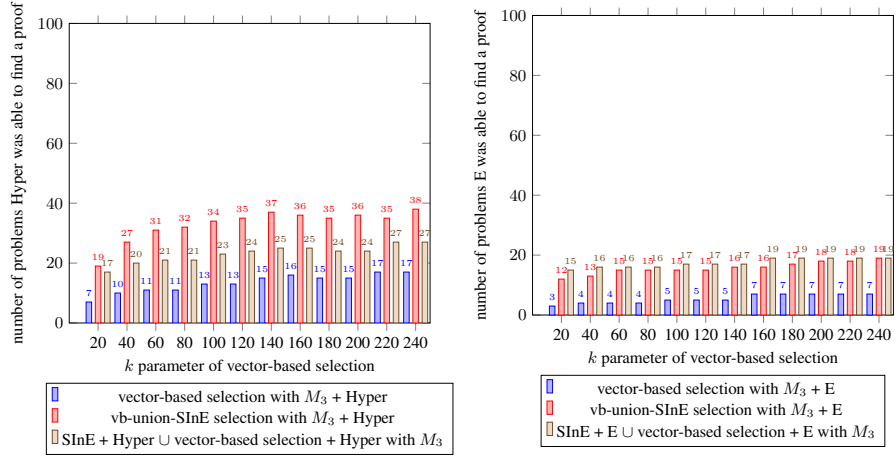
\begin{figure}[H]
\begin{subfigure}{.5\textwidth}
\begin{tikzpicture}[scale=0.7]
\begin{axis} [xtick={20,40,60,80,100,120,140,160,180,200,220,240},xticklabels={20,40,60,80,100,120,140,160,180,200,220,240},ybar, xmin=20,	bar width=3pt,
ymin=0,ymax=100, xmin=0,xmax=250,xlabel = $k$ parameter of vector-based selection, ylabel=  number of problems Hyper was able to find a proof, legend style={at={(0.5,-0.2)},anchor=north},nodes near coords, point meta=y, every node near coord/.append style={font=\tiny},legend style={cells={align=left}}]
    \addplot table [x=a, y=vb, col sep=comma] {data_hyper_m3.csv};
    \addplot table [x=a, y=sinevb, col sep=comma] {data_hyper_m3.csv};
    \addplot table [x=a, y=vereinigung, col sep=comma] {data_hyper_m3.csv};
    \legend{vector-based selection with $M_3$ + Hyper, \vbsine selection with $M_3$ + Hyper, SInE + Hyper $\cup$ vector-based selection + Hyper with $M_3$}  
\end{axis}
\end{tikzpicture}
\end{subfigure}
\begin{subfigure}{.5\textwidth}
\begin{tikzpicture}[scale=0.7]
\begin{axis} [xtick={20,40,60,80,100,120,140,160,180,200,220,240},xticklabels={20,40,60,80,100,120,140,160,180,200,220,240},ybar, xmin=20,	bar width=3pt,
ymin=0,ymax=100, xmin=0,xmax=250,xlabel = $k$ parameter of vector-based selection, ylabel=  number of problems E was able to find a proof, legend style={at={(0.5,-0.2)},anchor=north},nodes near coords, point meta=y, every node near coord/.append style={font=\tiny},legend style={cells={align=left}}]
    \addplot table [x=a, y=vb, col sep=comma] {data_eprover_m3.csv};
    \addplot table [x=a, y=sinevb, col sep=comma] {data_eprover_m3.csv};
    \addplot table [x=a, y=vereinigung, col sep=comma] {data_eprover_m3.csv};
    \legend{vector-based selection  with $M_3$ + E, \vbsine selection  with $M_3$ + E, SInE + E $\cup$ vector-based selection + E  with $M_3$}  
\end{axis}
\end{tikzpicture}
\end{subfigure}
\caption[]{Results of Hyper and E on 1,000 randomly selected Adimen SUMO problems that the respective prover could not solve without selection. The experiments use mapping$M_3$ and three 
different selections followed by runs of the two provers: vector-based selection, \vbsine and the afore described SInE + prover $\cup$ vector-based selection + prover.
For each setting, we provide the number of proofs found by the respective prover. 
Please note: To increase readability, we have set the maximum value of the y-axis to 100 which should be taken into account when comparing with diagrams from the previous section.
}
  \label{fig:vb_mapping_M3}
\end{figure}

\end{document}